\documentclass{article}
\usepackage[utf8]{inputenc}

\title{RapidRead: Global Deployment of State-of-the-art Radiology AI for a Large Veterinary Teleradiology Practice}

 \author{Michael Fitzke$^1$,  Conrad Stack$^1$, Andre Dourson$^1$, \\ Rodrigo M. B. Santana$^1$ Diane Wilson$^2$, Lisa Ziemer$^2$, \\  Arjun Soin$^3$, Matthew P. Lungren$^3$, Paul Fisher$^2$, Mark Parkinson$^1$}
\date{
    $^1$Mars Digital Technologies\\%
  $^2$Antech Imaging Services\\
$^3$Stanford University, Center for Artificial Intelligence in Medicine \& Imaging\\[2ex]}

\usepackage{multirow}
\usepackage{fancyhdr}
\usepackage{graphicx}
\usepackage{grffile}
\usepackage{amsmath}
\usepackage{amssymb}
\usepackage{hyperref}
\usepackage{lscape} 
\usepackage{geometry}
\usepackage{mathrsfs}
\usepackage{subfig}
\usepackage{booktabs}
\usepackage[anticlockwise,figuresright]{rotating}

\usepackage[numbers]{natbib}

\begin{document}

\maketitle

\begin{abstract}
This work describes the development and real-world deployment of a deep learning-based AI system for evaluating canine and feline radiographs across a broad range of findings and abnormalities. We describe a new semi-supervised learning approach that combines NLP-derived labels with self-supervised training leveraging more than 2.5 million x-ray images. Finally we describe the clinical deployment of the model including system architecture, real-time performance evaluation and data drift detection.

\end{abstract}

\section{Introduction}

Radiographic imaging is the most common clinical imaging modality in the world and important for clinical evaluation of patients both in human and veterinary medicine. The global veterinary imaging market size was valued at 2.01 billion in 2018 and increase in utilization for veterinary diagnostics is expected to be driven largely by rising demand for pet insurance and growing animal healthcare expenditure, increasing companion animal population, and growth in the number of veterinary practitioners globally \cite{noauthor_undated-ow}. Currently, while medical imaging use increases, the shortage of veterinary radiologists remains a critical problem globally as less than 5\% of practicing veterinarians have formal training in imaging interpretation \cite{noauthor_undated-ow}. Further challenges in veterinary radiology arise chiefly due to the wide variety of species, breeds, size, as well as inconsistent positioning and poor standardization, all of which contribute to misinterpretation and degraded clinical care \cite{Bruno2015-yp,Berlin2007-vw,Alexander2010-ny,Gatt2003-gj,Reese2011-ph, Waite2017-pe}. 

Recent advancements in machine learning in medical imaging applications have demonstrated expert-level performance for automated diagnoses across a variety of modalities and diseases in human radiology. Similar applications have been described in veterinary radiology, yet suffer from limitations stemming from dataset size, lack of multi-site data informing generalizability, and chiefly for the field of deep learning in medical imaging broadly, lack experimental data with real-world constraints via deployment of automated systems in practice which vastly limit clinical translation of these promising technologies.  Ultimately, to realize the potential for modern supervised deep learning in automated medical image diagnosis, large real-world, multi-center, multi-class datasets with human-expert annotations and consensus benchmarks are needed; further a framework for model deployment in practice at scale is ideal to facilitate translation of these technologies in medical imaging practice broadly \cite{Driver2020-ws}.

The purpose of this study is to explore state-of-the-art computer vision techniques leveraging a large multi-center dataset consisting of more than 2.5M veterinary radiographs, human expert supervision, multi-class end-to-end diagnostic tasks, and describe model best model performance in real-time deployment for a high-volume global teleradiology practice. Since the clinical workflows we aim to model change dynamically, this study also outlines data drift detection analysis that is tailored to a vision-based use case and presents a viable approach towards monitoring of large models deployed across multiple institutions, without requiring immediate access to expert-annotatated labels at inference time. 

This work could potentially serve as a vital framework for accurate clinical machine vision applications in radiology at scale; addressing critical limitations in the use of this technology to advance the health of pets while also providing insights for medical imaging human medicine \cite{Irvin2019-ob,Bien2018-ri, Larson2018-ts, Huang2020-fd, Rajpurkar2020-oz, Rajpurkar2017-ft, Dunnmon2019-fm, Rajpurkar2018-bh}.

\section{Data}

\subsection{Images}

\begin{figure}
\includegraphics[width=1.0\textwidth]{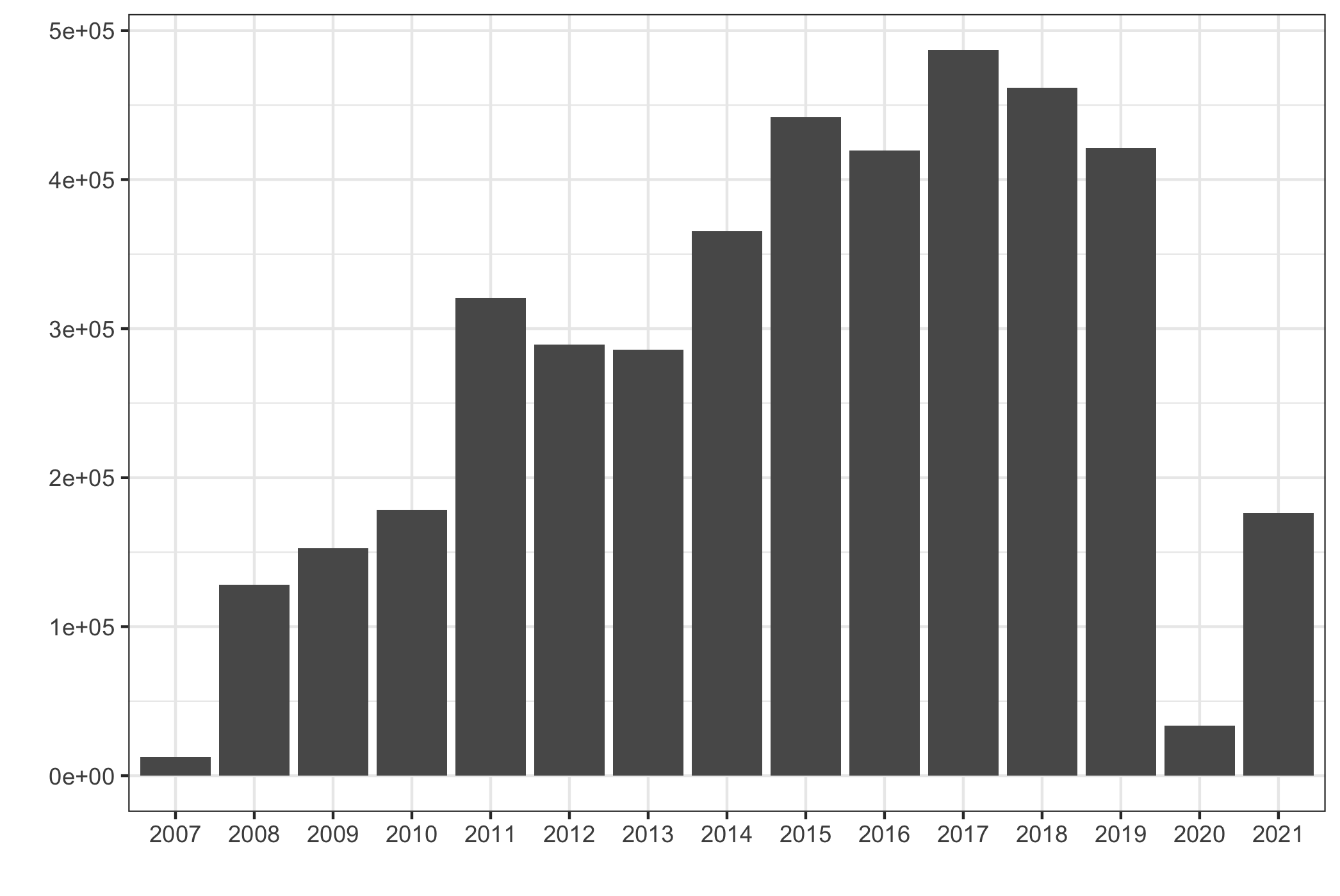}
\caption{Number of images provided, by year}
\label{fig:imagesperyear}
\centering
\end{figure}

Over 3.9 million veterinary radiographs were available for modeling.  The majority of these radiographs had previously been archived as lossy (quality 89) JPEG images, down-sampled to a fixed width of 1024 pixels (px) (`set 1' in Table \ref{tab:imagesummary}).  Of the remaining radiographs most were provided as (lossless) PNG images that were down-sampled so the smaller dimension (width or height) was 1024 px (`set 2' in Table \ref{tab:imagesummary}). The final subset of images were collected as part of the current clinical workflow (Figure \ref{fig:infra}), where DICOM images were down-sampled to a fixed height of 512 px, then converted to PNG (`silver' in Table \ref{tab:imagesummary}).  In all cases, the downsampling process preserved original aspect ratios.  All images were provided by Antech Imaging Services (AIS) along with a subset of the original DICOM tags as metadata. All images / studies cover real clinical cases from various client hospitals and clinics ($N > 3500$) received by AIS within the last 14 years (Figure \ref{fig:imagesperyear}).

A number of filtering steps were applied prior to modeling.  First, duplicated and low-complexity images were removed using \textit{imagemagick}, open-source software for displaying, creating, converting, modifying, and editing images. Second, imaging artefacts, and views / body parts irrelevant to our set of findings were filtered out using a CNN model trained for this purpose. Studies with more than 10 images were also excluded. Approximately $2.7$ million images were left after filtering, representing over $725,000$ distinct patients.

\begin{table}
\caption{\label{tab:imagesummary} Summary of x-ray image data}
\begin{tabular}{lllll}
\toprule
{} &         Set 1 &            Set 2 &       Silver &   (Total) \\
\midrule
No. of images before filtering &       3318579 &           455077 &       186563 &   3960219 \\
No. of images after filtering  &       2202007 &           344050 &       186563 &   2732620 \\
(\% Filtered out)               &      0.336461 &         0.243974 &            0 &  0.309983 \\
Image format(s)                &          JPEG &              PNG &          PNG &       \\
Range of image widths     &          1024 &  [1024 - 40345]  &  [71 - 2516] &       \\
Range of image heights        &  [258 - 3068] &   [1024 - 35328] &          512 &       \\
\bottomrule
\end{tabular}
\end{table}

\subsection{Annotation / Labeling}

Images were annotated for the presence of 41 different radiological observations (Table \ref{tab:labels}).  For the majority of images (pre-2020 studies) labels were extracted from corresponding (study-wise) radiological reports using an automated, natural language processing (NLP)-based algorithm.  The radiology reports summarize all the images in a particular study, and were written by over $2000$ different board-certified veterinary radiologists.  Images from the most recent studies (years 2020-2021; `silver' in Table \ref{tab:imagesummary}) were individually labeled by a veterinary radiologist, immediately after they finished evaluating the study.  

In order to assess the trained model's accuracy and inter-annotator variability, a small number of images (N=$615$) were randomly selected from the `silver' set to be labeled by twelve additional radiologists.  These data were not used in training or validation.  To produce ground truth labels for ROC and PR analyses the labels for each image were aggregated by a majority-rules vote (i.e., a majority of the 12 radiologists indicated the finding was present).  Point estimates of the False Positive Rate (FPR) and Sensitivity for a particular radiologist were calculated by comparing their labels to the majority-rules vote of the 11 others. 

Automated extraction of labels from the ``Findings" section of radiology reports was done using a modified version of the rules-based labeling software, CheXpert \citep{Irvin2019-ob, Peng2018-ot}. We extended the number of labels used be Chexpert to 41, and adapted it's syntactic patterns and key-phrases to reflect the language found in our report data, and veterinary medicine more generally.

While these reports contain observations from all images in a study, they never explicitly link these observations with specific image files. Therefore, the per-study labels extracted from a report were initially applied to all images in the corresponding study, and then masked using a set of expert-provided rules.  These rules ensure that labels are only applied to images that show the corresponding body parts (e.g., `Cardiomegaly' labels are removed from images of the pelvis).

\begin{table}
\centering
\caption{\label{tab:labels} Radiological labels}
\begin{tabular}{ll}
\toprule
Anatomical Grouping & Observation \\
\midrule
\multirow{9}{*}{Cardiovascular}  &                                       Cardiomegaly \\
               &                            Left Atrial Enlargement \\
               &                       Left Ventricular Enlargement \\
               &                      Right Ventricular Enlargement \\
               &                           Right Atrial Enlargement \\
               &                  Main Pulmonary Artery Enlargement \\
               &                                 Aortic Abnormality \\
               &                             Heart Base Mass Effect \\
               &                                        Microcardia \\
\cline{1-2}
\multirow{6}{*}{Pulmonary Structures}  &         Bronchial pattern \\
               &  Interstitial Unstructured \\
               &                                 Pulmonary Alveolar \\
               &       Pulmonary Interstitial - Nodule (Under 1 cm) \\
               &                                 Pulmonary Vascular \\
               &                         Pulmonary Mass (Over 1 cm) \\
\cline{1-2}
\multirow{4}{*}{Pleural Space} &        Sign(s) of Pleural Effusion \\
               &                                Pleural Mass Effect \\
               &                                       Pneumothorax \\
               &                        Thin Pleural Fissure Lines \\
\cline{1-2}
\multirow{6}{*}{Mediastinal Structures} &       Esophagal Dilation \\
               &                   Intrathoracic Tracheal Narrowing \\
               &                                 Tracheal Deviation \\
               &                                   Mediastinal Mass \\
               &                               Mediastinal Widening \\
               &           Mediastinal Lymph Node Enlargement \\
\cline{1-2}
\multirow{16}{*}{Extra Thoracic} &                      Spondylosis \\
               &                                  Liver Abnormality \\
               &                                    Abdominal mass \\
               &                        Intervertebral Disc Disease \\
               &                           Gastric Foreign Material \\
               &             Cervical Tracheal Narrowing or Opacity \\
               &                         Degenerative Joint Disease \\
               &                           Decreased serosal detail \\
               &                                 Gastric Distention \\
               &                             Aggressive Bone Lesion \\
               &                           Fracture and/or Luxation \\
               &                                       Splenomegaly \\
               &                        Gastric Dilatation Volvulus \\
               &                                Subcutaneous Nodule \\
               &                                  Subcutaneous Mass \\
               &                     Fat Opacity Mass (e.g. lipoma) \\
\bottomrule
\end{tabular}

\end{table}

\section{Methods}

\subsection{Noisy Radiology Student}
Using our data sets of individually labeled data $(x_1, y_1), \ldots, (x_n, y_n)$ and data labeled by the Chexpert labeler \cite{Irvin2019-ob} adapted to veterinary radiology reports  $(\tilde{x}_1, \tilde{y}_1), \ldots, (\tilde{x}_m, \tilde{y}_m)$. For each finding $\tilde{y}_i$ has as entry either $0$ (negative), $1$ (positive) or  $u$(uncertain) and is determined on the level of the radiology report instead on the level of the individual image. Hence label noise is present on the latter data set. 

To combine both data sets into a single mode we combine the distillation approaches of  \cite{li2017learning} and \cite{xie2020self} and adapt them to our multi-label use case. We first train a teacher model $\theta_*^t$ with the individually labeled data set and add noise to the training process:

\begin{align*}
    \frac{1}{n} \sum_{i=1}^n l(y_i, f^{noise}(x_i, \theta^t)).
\end{align*}

After that  we use our teacher model to infer soft pseudo labels for the images $\tilde{x}_1, \ldots \tilde{x}_m$. In the inference step we do not use any noise:

\begin{align*}
    \hat{y}_i = f(x_i, \theta_*^t), \forall i \in (1, \ldots, m).
\end{align*}

We then combine the soft pseudo labels with the NLP derived labels using the rul:

\begin{align*}
    \bar{y}_i = \begin{cases}\lambda \hat{y}_i +  (1-\lambda) 0.5& \text{if } \tilde{y}= u\\
\lambda \hat{y}_i +  (1-\lambda) \tilde{y}_i & \text{otherwise} 
    \end{cases}
    , \forall i \in (1, \ldots, m).
\end{align*}

With those derived labels we train an equal-or-larger student model $\theta_*^s$ with noise added to the student: 

\begin{align*}
    \frac{1}{n} \sum_{i=1}^n l(y_i, f^{noise}(x_i, \theta^s)) + \frac{1}{m} \sum_{i=1}^m l(\bar{y}_i, f^{noise}(\tilde{x}_i, \theta^s))  .
\end{align*}

In our production setting we set the student model as new teacher and re-do the process without the first step whenever we have collected a significant amount of new single-labeled data. 
\par
\paragraph{Training details.} We train different artificial neural network architectures, specifically Densenet-121 \cite{huang2017densely}, Inception-v4 \cite{szegedy2016inception} and Models from the EfficientNet-Family \cite{tan2019efficientnet}. For each model we maximize the number of images that fit into memory to determine batch-size which lead to batch-sizes between 32 and 256. We use different image-input-sizes (ranging from $224\times224$ to $456\times456$) and train both with reshaping the original images to the input dimension as well as zero-padding the image to a square and then resizing to keep the ratios of the original image intact. 
We perform a multitude of image augmentation techniques when training such as in \cite{cubuk2020randaugment}. All our models were pre-trained on Imagenet \cite{deng2009imagenet} and are trained for a maximum of 30 epochs with early stopping if the validation loss is not decreasing two epochs in a row. 

\paragraph{Probability Calibration}
In order to calibrate the probability for each finding, we apply the piecewise linear transformation \cite{cohen2019chester} : 

\begin{align*}
    t_{\phi, opt_\phi}(x) = 
    \begin{cases}
    \frac{x}{2\cdot opt_\phi},& \text{if } x\leq opt_\phi\\
    1 -  \frac{1-x}{2\cdot(1-opt_\phi)},              & \text{else}
\end{cases}
\end{align*}

for all findings $\phi$. We set $opt_\phi$ to optimize Youden's J Statistic \cite{youden1950index} on an independent validation set.

\subsection{Drift Analysis}
 In veterinary radiology at the scale described in this work we could well have multiple factors leading to a difference between data distributions seen pre and post model deployment. These factors include changes in breeds, different radiology equipment or differences in clinical practice in different regions. Such changes can have implications on performance metrics and necessitate checking whether assumptions made during the development of models still hold while models are in production \cite{datasetshift,conceptdrift,rabanser2019failing}. 
 
At time of development of the system the images $X_1, \ldots X_n$ and findings per image $Y_1, \ldots, Y_n$ can be viewed as draws from the joint distribution $P_{dev}(X,Y)$. In a real world production scenario we are presented with images and findings from a distribution $P_{prod}(X,Y)$. In this work we specifically investigate covariate shift, i.e a change in the marginal distribution: 

\begin{align*}
    P_{dev}(Y|X) &= P_{prod}(Y|X) \\
    P_{dev}(X) &\neq P_{prod}(X)
\end{align*}

and its influence on model performance. Qualitatively, covariate shift refers to the change in the distribution of the input features present in the training and the test data. It occurs when some previously infrequent feature vectors become more frequent, and vice versa, while the relationship between the feature and the target classes remain the same \cite{conceptdrift, rabanser2019failing}. 

To detect covariate shift we train an autoencoder $f_A(\cdot, \Theta)$ at dev-time, i.e. try minimizing $\int (X - f_A(X|\Theta))^2 dP_{dev}$ with regards to $\Theta$.  Training was done using the Alibi-Detect software \cite{alibi-detect}. We then use the trained autoencoder to reconstruct images at production time and analyze the reconstruction errors to gauge drift. 

 While supervised drift detection is achieved by monitoring a performance metric (AUROC, F1, PR scores) of interest over time, it requires access to an abundance of labels at inference time to quantify a change in the system performance - a requirement that is cost-prohibitive and ideally requires expert radiologist annotators. Our approach does not make any assumptions about the model that has been deployed, and requires access to training data used to build the model as well as production data ingested by the model post-deployment. We build a machine-learned representation (autoencoder) of the training dataset and use this representation to reconstruct data that is presented to the model in the real world. By comparing reconstruction errors, we effectively detect changes in the input (X-Ray image data) seen in production.

\subsection{Towards Scaling Laws for X-Ray Images}
Academic datasets of X-Ray images that are categorized as large (e.g. \cite{Irvin2019-ob}, \cite{johnson2019mimic}) have under 500,000 individual images while our dataset has over 2 Million NLP-labeled images and 180.000 individually labeled images by board certified radiologists. This allows us to start investigating questions of data scaling and related questions on model capabilities.

\subsection{Ensembling}
Our final model (RapidReadNet) is an ensemble of individual, calibrated deep neural networks. We found that averaging the outputs leads to slightly better metrics than voting. We use the $8$ best models based on our validation set and then used the best subset method to determine the optimal ensemble. Surprisingly, the best subset was the full set of $8$ models, i.e. models that were under-performing against the rest of the ensemble still helped overall predictions. 

\subsection{Study level fusion}
To determine the findings per study we merge individual findings per image by choosing the disease class mapping to the $argmax$ value per finding over all images.

\section{Results}

\subsection{Results on a highly curated data set  }

\begin{figure}
\includegraphics[width=1.0\textwidth]{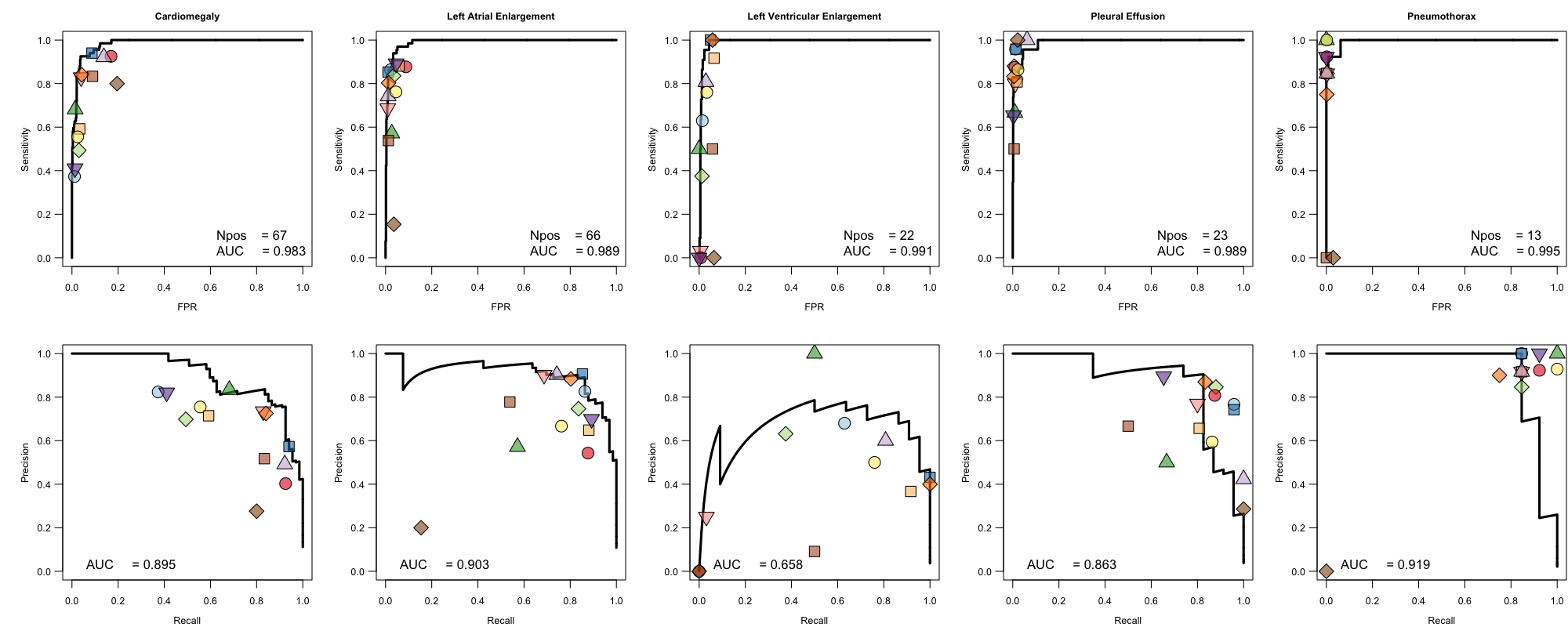}
\caption{ROC and PR curves for Cardiovascular and Pleural Space findings}
\label{fig:plat_cardio}
\centering
\end{figure}

\begin{figure}
\includegraphics[width=1.0\textwidth]{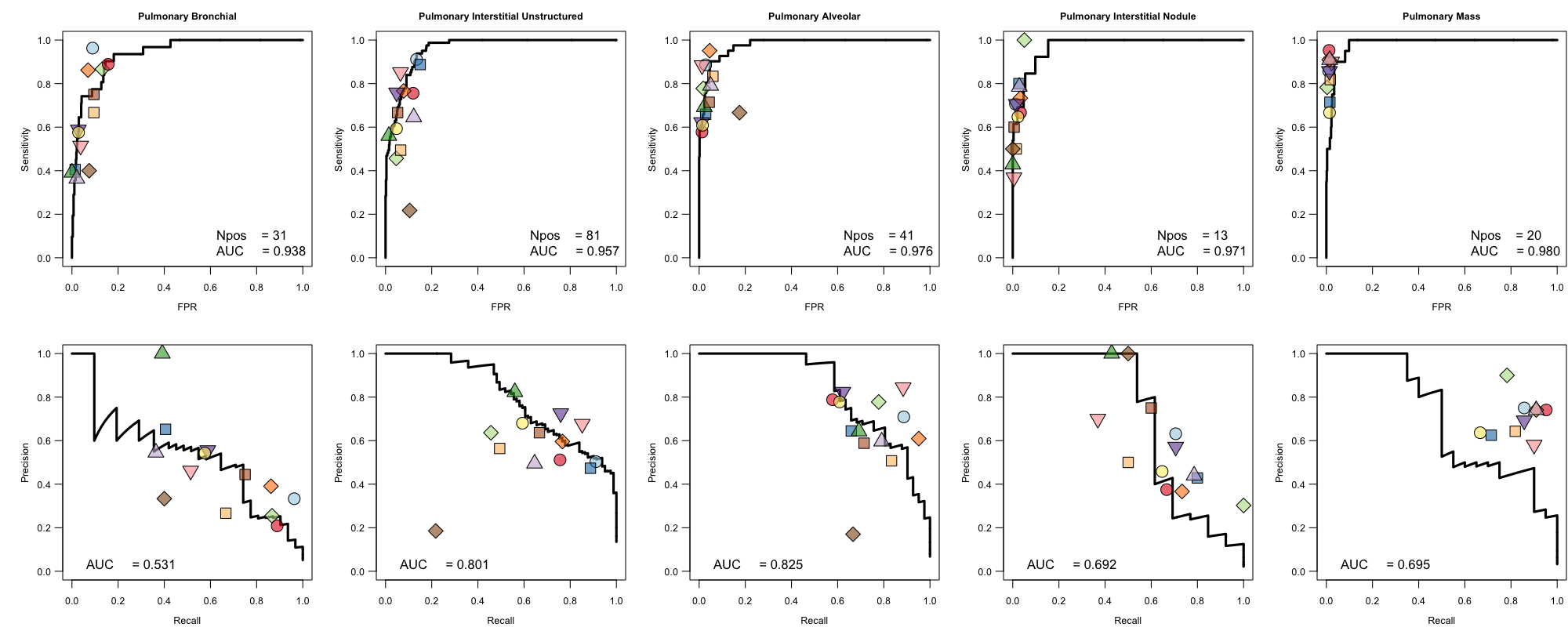}
\caption{ROC and PR curves for Pulmonary findings}
\label{fig:plat_pulmonary}
\centering
\end{figure}

\begin{figure}
\includegraphics[width=1.0\textwidth]{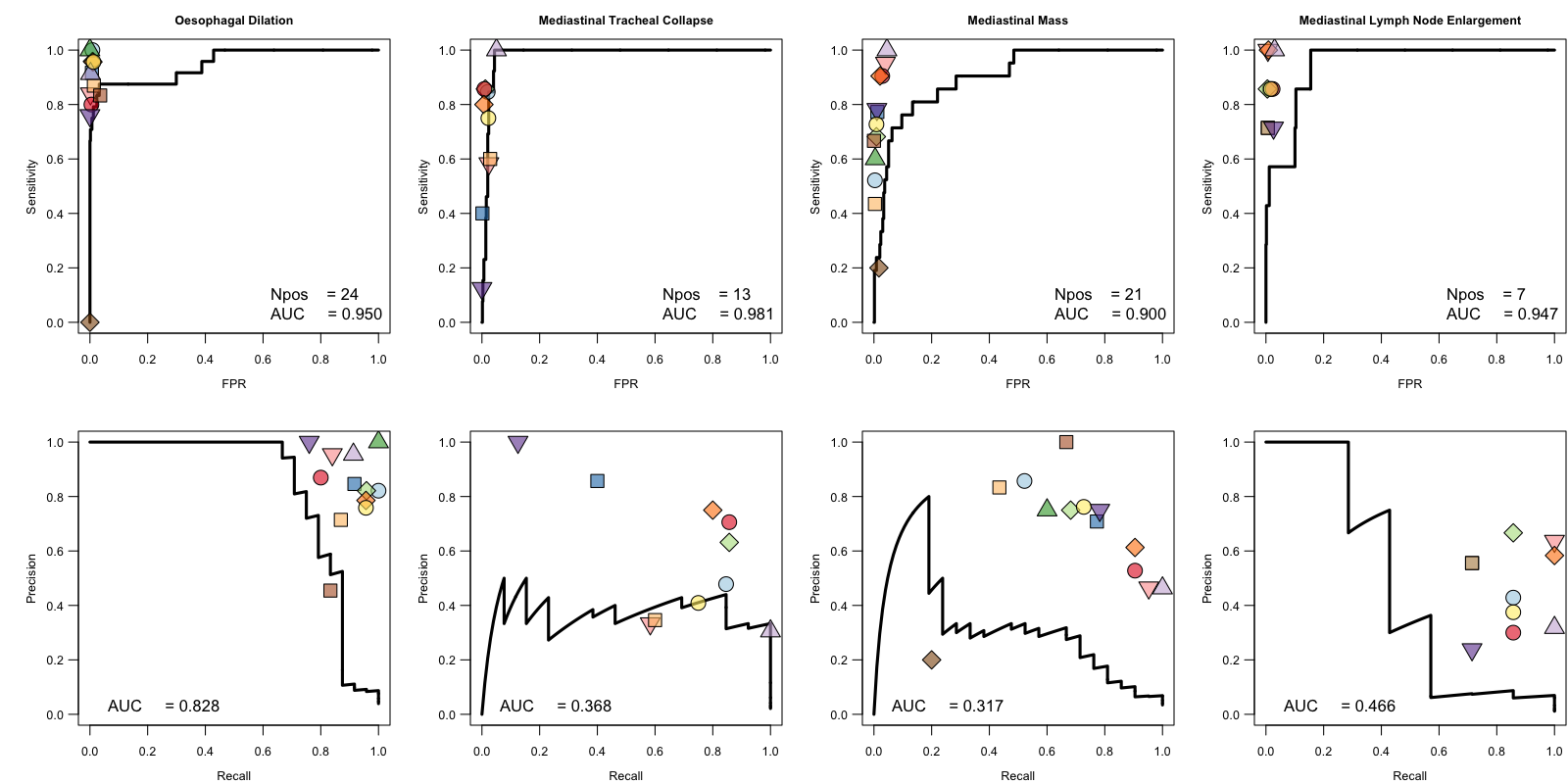}
\caption{ROC and PR curves for Mediastinal findings}
\label{fig:plat_mediastinal}
\centering
\end{figure}

\begin{figure}
\includegraphics[width=1.0\textwidth]{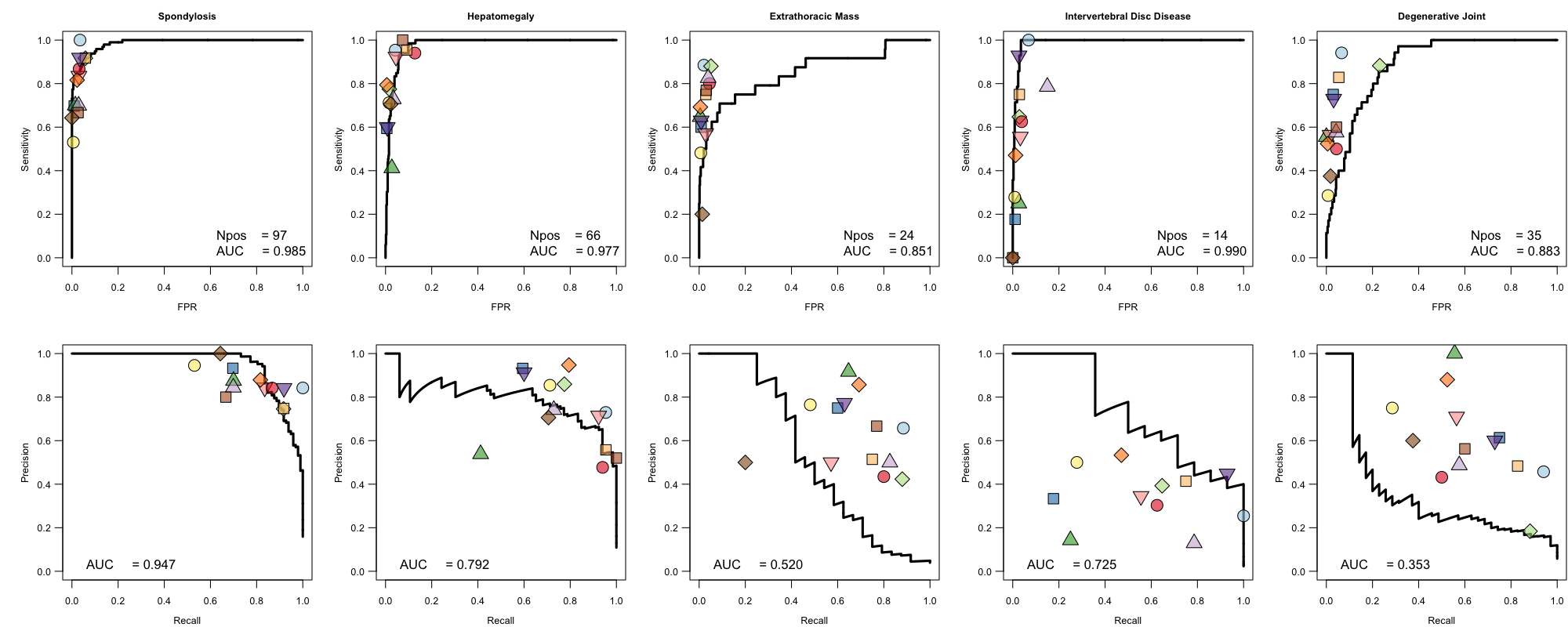}
\caption{ROC and PR curves for Extrathoracic findings (part 1)}
\label{fig:plat_et1}
\centering
\end{figure}

\begin{figure}
\includegraphics[width=1.0\textwidth]{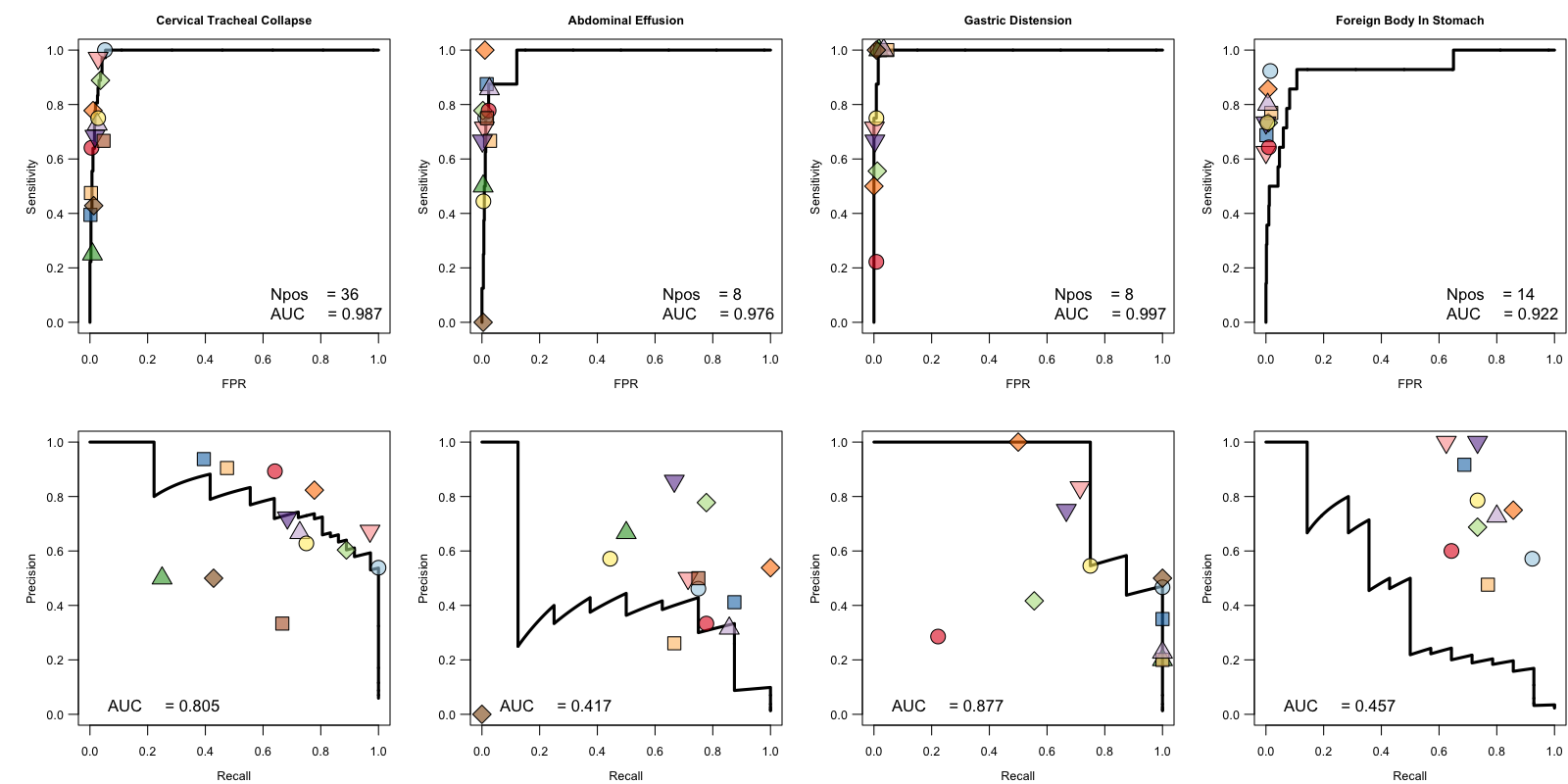}
\caption{ROC and PR curves for Extrathoracic findings (part 2)}
\label{fig:plat_et2}
\centering
\end{figure}

Figures \ref{fig:plat_cardio}, \ref{fig:plat_pulmonary}, \ref{fig:plat_mediastinal}, \ref{fig:plat_et1}, \ref{fig:plat_et2} show the result ROC and PR analysis results for the set of $613$ images labeled by multiple radiologists, comparing model predictions against the radiologists' labels. Each figure shows ROC (top) and PR (bottom) curves per-finding, and point estimates of each individual radiologist's FPR, Precision, and Recall (Sensitivity).  Findings with fewer than five positive labels were not analyzed. 

For a majority of the findings, model accuracy (AUROC) is comparable to the accuracy (AUROC) of individual radiologists.

\subsection{Longitudinal drift analysis}

\begin{figure}
\includegraphics[width=1.0\textwidth]{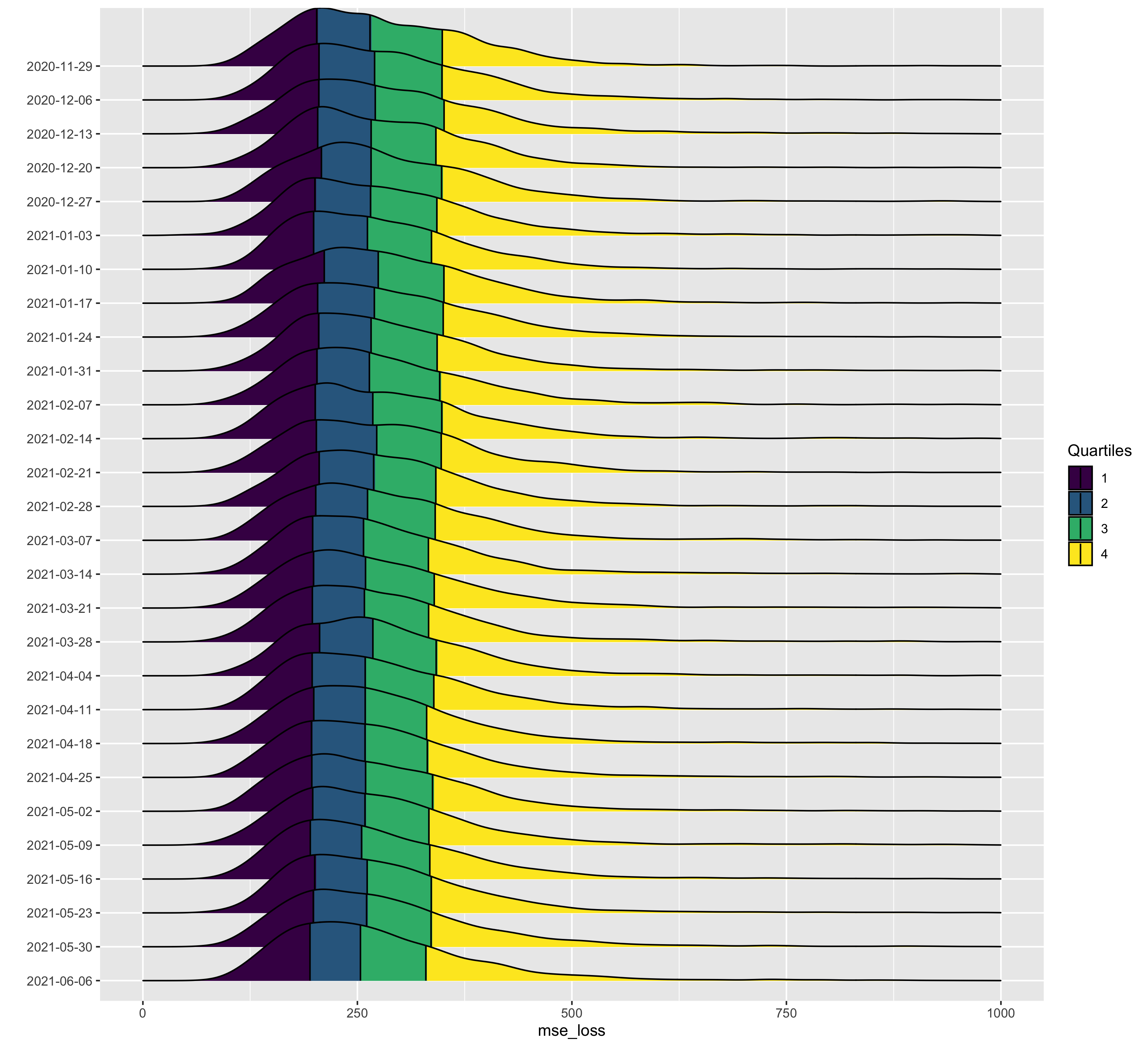}
\caption{Weekly reconstruction error}
\label{fig:errorovertime}
\centering
\end{figure}

An autoencoder was trained using all archived images (sets 1 and 2 in Table \ref{tab:imagesummary}), and then applied to the images from subsequent studies, checking the L2 reconstruction error of each.  Figure \ref{fig:errorovertime} shows the distribution and quantiles of L2 errors, grouped by week.  

We observe little-to-no difference between reconstruction errors; indicating that input data has remained distributionally consistent, and further suggesting that the the model is robust to data from new clients.We also show that a convolutional autoencoder is capable of quantifying differences in X-Ray image datasets on the basis of a reconstruction error. Formally, $P(X)$ and $P(Y|X)$ exhibit almost no change, and we expect the model’s performance on new data to remain comparable to testing baselines:

\begin{align*}
    P_{dev}(Y|X) &\approx P_{prod}(Y|X) \\
    P_{dev}(X) &\approx P_{prod}(X)
\end{align*}

\begin{figure}
\includegraphics[width=1.0\textwidth]{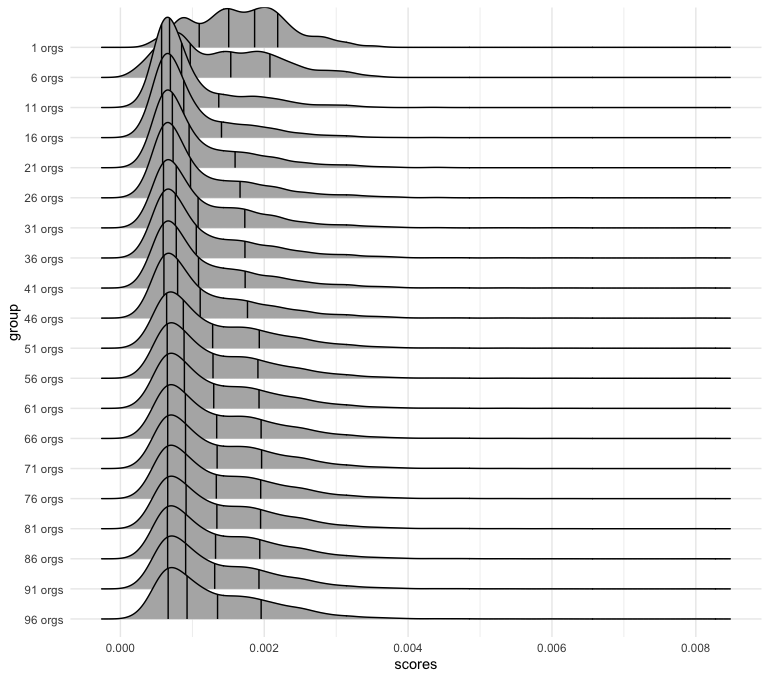}
\caption{Loss distribution vs. number of organizations}
\label{fig:errorbynorgs}
\centering
\end{figure}

This lack of perceived drift can be attributed, in part, the high diversity of organizations and animals represented in the training data.  Figure \ref{fig:errorbynorgs} shows the distribution of reconstruction error as a function of the number of organizations represented.  Data from a small number of organizations (1,6,..,16 orgs) does not provide a good representation of the total diversity of the image data.  

\subsection{Data Scaling}

We observed a positive relationship between data size and model performance on an independent, hand labeled test set. We trained an Efficient-Net-b5 \cite{tan2019efficientnet} on differently sized subsets of our data and tested the resulting models on the same test set. Results can be observed in Table \ref{tab:scale} and suggest the possibility of further performance gains with data scaling.

\begin{table}[h]
\centering
\begin{tabular}{lll}
\toprule
                & Roc-AUC  & PR-AUC      \\
\midrule
20,000 images & 0.8765   & 0.3140     \\
50,000 images  & 0.8974 & 0.3706 \\ 
100,000 images & 0.907  & 0.393 \\
500,000 images & 0.912 &  0.404 \\
1,000,000 images & 0.916& 0.433\\
1,500,000 images & 0.920& 0.443 \\
2,000,000 images & 0.922 & 0.477 \\
2,500,000 images & 0.926 & 0.488 \\
                   &          &   \\
\bottomrule
\end{tabular}

\caption{\label{tab:scale} Metrics of our models on 30477 unseen test data points against ground truth of one board certified radiologist. All data was labeled in clinical production setting. }
\end{table}

\subsection{Study level results}

Results are shown in Table \ref{tab:studywise}. Additional information can be found in the supplemental file, \path{ROC-report_silver_2021-09-13_STUDYWISE.pdf}, including ROC plots broken down by species (ie `canine', `feline').

\begin{table}
\centering
\caption{Study-wise ROC results for each finding.  The Npositive column lists the number of studies with at least one positive label (out of 9311 total studies). Area under the Receiver-Operating Curve (AUROC), False positive rate (FPR), and Sensitivity were not calculated for findings with fewer than 10 positive instances (Npositive). }
\label{tab:studywise}
\begin{tabular}{lllll}
\toprule
{} & Npositive & AUROC & FPR @ 0.6 & Sensi @ 0.6 \\
\midrule
Cardiomegaly                               &                                          915 & 0.986 &     0.025 &             0.923 \\
Left Atrial Enlargement                    &                                          738 & 0.988 &     0.016 &             0.912 \\
Left Ventricular Enlargement               &                                          479 & 0.994 &     0.005 &             0.939 \\
Right Ventricular Enlargement              &                                          157 & 0.971 &     0.005 &             0.694 \\
Right Atrial Enlargement                   &                                           90 & 0.984 &     0.003 &             0.767 \\
Main Pulmory Artery Enlargement            &                                           13 & 0.826 &         0 &             0.308 \\
Aortic Abnormality                         &                                           20 & 0.687 &         0 &                 0 \\
Heart Base Mass Effect                     &                                            1 &  n/a &      n/a &              n/a \\
Spondylosis                                &                                         2133 & 0.993 &     0.010 &             0.962 \\
Liver Abnormality                          &                                         1294 & 0.969 &     0.041 &             0.934 \\
Ex. Thoracic or abdomil mass               &                                          502 & 0.960 &     0.017 &             0.843 \\
Sign(s) of IVDD                            &                                          810 & 0.951 &     0.018 &             0.806 \\
Gastric Foreign Material                   &                                          357 & 0.903 &     0.011 &             0.650 \\
Cervical Tracheal narrowing or Opacity     &                                          632 & 0.983 &     0.015 &             0.897 \\
Degenerative Joint Disease                 &                                          900 & 0.945 &     0.037 &             0.810 \\
Decreased serosal detail                   &                                          410 & 0.964 &     0.010 &             0.793 \\
Gastric Distention                         &                                          500 & 0.975 &     0.012 &             0.912 \\
Aggressive Bone Lesion                     &                                           41 & 0.848 &     0.000 &             0.049 \\
Fracture and/or Luxation                   &                                          101 & 0.753 &         0 &                 0 \\
Esophagal Dilation                         &                                          213 & 0.986 &     0.010 &             0.897 \\
Intrathoracic Tracheal rrowing             &                                          323 & 0.974 &     0.010 &             0.796 \\
Tracheal Deviation                         &                                          252 & 0.985 &     0.003 &             0.825 \\
Mediastil Mass                             &                                           99 & 0.965 &     0.007 &             0.778 \\
Mediastil Lymph Node Enlargement (any)     &                                           56 & 0.911 &     0.001 &             0.536 \\
Sign(s) of Pleural Effusion                &                                          238 & 0.985 &     0.010 &             0.924 \\
Pleural Mass Effect                        &                                            4 &  n/a &      n/a &              n/a \\
Pneumothorax                               &                                           44 & 0.990 &     0.001 &             0.750 \\
Bronchial pattern                          &                                         1745 & 0.959 &     0.057 &             0.890 \\
Interstitial Unstructured                  &                                         1300 & 0.974 &     0.026 &             0.906 \\
Pulmory Alveolar                           &                                          798 & 0.986 &     0.018 &             0.910 \\
Pulmory Interstitial - Nodule (Under 1 cm) &                                          218 & 0.934 &     0.010 &             0.638 \\
Pulmory Vascular                           &                                          137 & 0.938 &     0.004 &             0.599 \\
Pulmory Mass (Over 1 cm)                   &                                          206 & 0.956 &     0.008 &             0.738 \\
Splenomegaly                               &                                          194 & 0.941 &     0.009 &             0.732 \\
Gastric Dilatation Volvulus                &                                           11 & 0.970 &     0.000 &             0.818 \\
Microcardia                                &                                           75 & 0.968 &     0.005 &             0.773 \\
Mediastil Widening                         &                                          131 & 0.962 &     0.004 &             0.748 \\
Pleural Fissure Lines                      &                                          579 & 0.981 &     0.038 &             0.957 \\
Subcutaneous Nodule                        &                                            9 &  n/a &      n/a &              n/a \\
Subcutaneous Mass                          &                                           77 & 0.950 &     0.001 &             0.506 \\
Fat Opacity Mass (e.g. lipoma)             &                                          253 & 0.984 &     0.006 &             0.893 \\
\bottomrule
\end{tabular}
\end{table}

\section{Deployment}
\paragraph{Deployment pipeline and workflow:} We now describe the system architecture of RapidReadNet. Our infrastructure pipeline is exclusively relying on micro-services (rest-APIs) deployed using docker containers \cite{docker}. Each container is using the FastApi \cite{FastApi} framework in order to deploy a rest API module, and each of them are serving as per best practices for such unique and specialized tasks. 

Given the amount of images processed each day (around 15,000+) we adopted an asynchronous processing approach for our production pipeline using a message broker. This was achieved using both Redis \cite{redis}, which is a NoSQL database component where we store each individual incoming requests, and use Redis Queue \cite{redisqueue} as a background processing mechanism to consume  each of the stored requests in parallel. 

More specifically, each request consists of an image and its corresponding metadata. Thus, incoming requests are processed by a component named Redis Queuer which takes care of caching the data locally and registering the request into the Redis DB's queue. Next, RQ workers are processing in parallel each of the requests stored in the queue and sending them to the model serving module called AI Orchestrator. 

All the predictions from our models are gathered in JSON format \cite{pezoa2016foundations} and are directly stored into a MongoDB \cite{mongodb} database (for long-term archiving) and a Redis JSON \cite{redisjson} database (for short term archiving). The short term storage (Redis JSON) allows us to aggregate results at the study-level along with inclusion of some contextualisation aspects.

Last but not least, for monitoring purposes, we deployed a Redis dashboard instance \cite{redisdashboard}, where each request can be tracked and re-processed if needed.

The micro-services are managed through docker compose \cite{compose} and organized using the following categories:

\begin{itemize}
\item Message broker: Pre-processing of incoming requests, monitoring and dispatch
\item Models serving: AI Orchestrator module, and individual models serving. All models are using Pytorch framework\cite{paszke2019pytorch}
\item Results and feedback loop storage: Aggregation and contextualization of models results at study-level, sending back of the results
\end{itemize}

\begin{figure}
\includegraphics[scale=0.3]{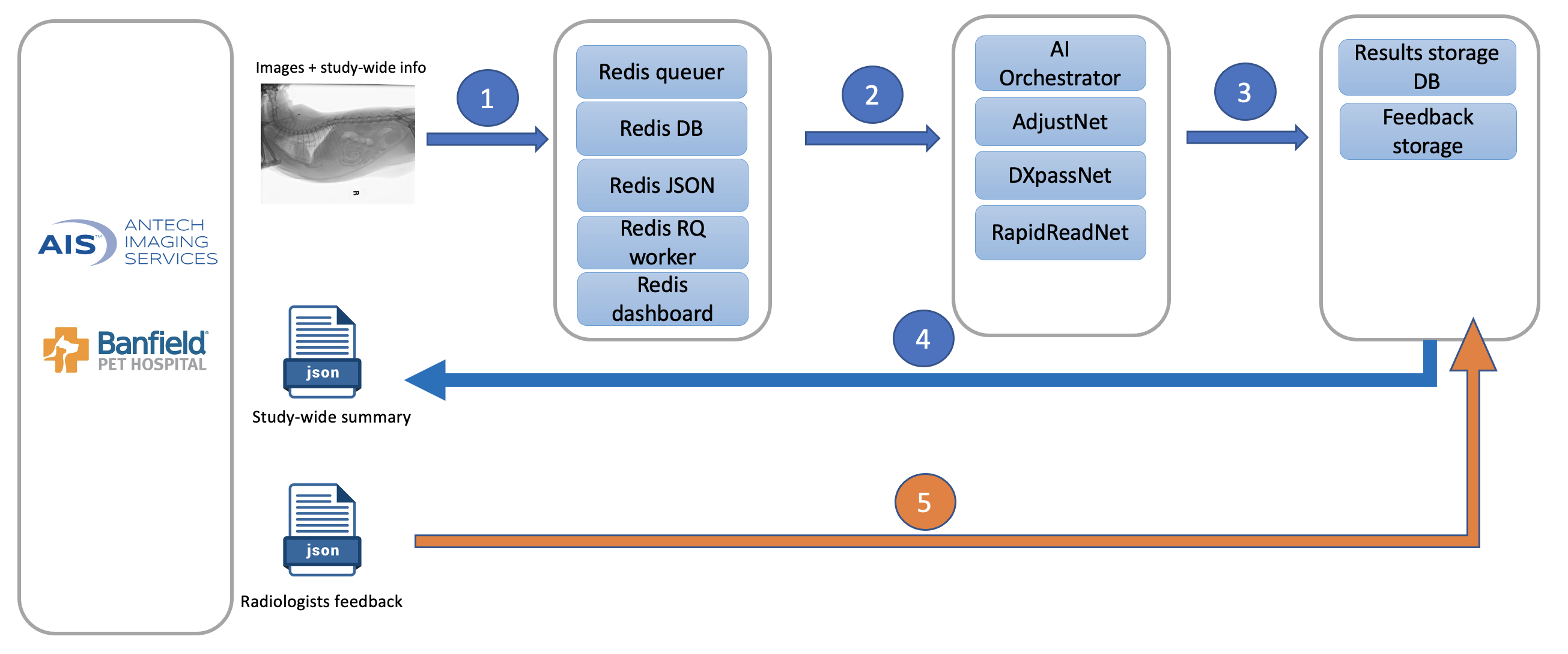}
\caption{Infrastructure of X-Ray system}
\label{fig:infra}
\centering
\end{figure}

An infrastructure diagram is depicted in Figure \ref{fig:infra}

\textbf{Models serving}

For serving models, the central element is the AI Orchestrator container, which is responsible to coordinate the execution of the inferences from our different AI modules. 

The first AI model from which predictions are collected is the \textit{AdjustNet}. This model checks the orientation of the radiograph image, and proceeds with adjustment if required (rotation, horizontal flipping, vertical flipping etc.). 

Then comes next the \textit{DxpassNet} which validates that the image presented to the model is appropriate for the diagnostic tasks provided by RapidReadNet, a multi-label classifier associated to 41 findings corresponding to various pathologies.

Lastly, a contextualization of the results is done using the study-wide metadata which were provided along with the image during the initial upload to the service. To achieve this aggregation at study level, a record of all study-wide image inferences is temporarily stored in the Redis JSON module, and we are using a rule-based expert system tool called CLIPS \cite{clips} to manage them from this contextualization. CLIPS (C Language Integrated Production System) was developed at NASA's Johnson Space Center in order to generate an AI rule-based system for space shuttle flight controllers. With that tool we were able to accommodate the rules applied against the outputs of our models and animal metadata in order to get context for the radiologist reports.
In order to productionize CLIPS we use a python library called clipspy \cite{clipspy} which interacts with C tools, while we save all the rules in a  mongoDb database in order to have a rule dynamism as new rules are created periodically.

\textbf{Results and feedback loop storage}

All records are stored in a mongoDb database in JSON format.

2. Embedded pre- and post- deployment infrastructure\par
3. Number of clinics, radiologist using system\par
4. Workflow for feedback on labels for radiologists (screenshot for structured labels)\par
5. Methods for adding in new labels as data is acquired - semi-supervised approach \par
6. Canary/shadow performance description of process\par

\section{Discussion}

The purpose of this study was to explore the development of a deep learning-based system to automate the detection of pre-defined clinical findings in canine and feline radiographs. We introduced a new method for data distillation that combines automated labeling with distillation and showed performance enhancements against the automated-labeling-only approach. We explored data scaling and its interaction with different models. We then look into the model's performance over time and compare it with input data drift over time. The input drift method uses a machine-learned representation of the image data to analyze whether subsequent X-Ray instances seen in production have drifted over time, and does not assume access to expert labels. Lastly we present the deployment process and the embedding of the model in a larger deep learning-based platform for X-ray image processing. This work has salient implications for the development, deployment and lifecycle management of deep learning-based systems in radiology practices. 

Human medical technological advancements have been driven by animal models for centuries; leading to the discovery of vitamins, hormones, antibiotics, transfusion, organ transplantation, vaccines, insulin, hemodialysis, chemotherapy, advanced diagnostic imaging techniques, and countless others. Animal medical research has saved millions of lives, human and non-human, and has transformed the world many times over.  The utilization of high-performance deep learning diagnosis systems at scale in veterinary care provide critical insights that can serve to address the gap in translation for these promising technologies in human and veterinary medical imaging diagnostics into clinical practice.

This work has a larger scope than any previous work in veterinary radiology. For example, a retrospective evaluation of a multi-class CNN trained with a dataset of only 3800 single-view lateral canine thoracic radiographs demonstrated modest performance on hold out data \cite{banzato2021automatic}. The lack of external validation, limited data size, narrow use case, and lack of deployment insights diminish real-world applicability. Another retrospective study was recently reported leveraging both canine and feline thoracic radiographs with two views (frontal and lateral) and three clinical labels in a dataset of 2800 images reported less than state-of-the-art performance compared to human radiology machine learning clinical models \cite{Arsomngern2019-sp} of comparable scale. The fact that the dataset presented in this paper is thousand orders of magnitude larger than those in aforementioned studies leads to significant gains in generalizability and robustness effects.




\paragraph{}
In conclusion, this work represents a first-of-its-kind, extensive analysis of the development and real-world application of deep learning methods in veterinary health. It shows promising potential towards using deeper architectures when scaling up the size of the data, as seen particularly in how the Noisy Radiologist Student boosts robustness in X-Ray image prediction. Furthermore, we analyze a large longitudinal dataset of medical imaging data for covariate shift and model performance and demonstrate the usefulness of clincal variability in helping avoid data drift. Lastly, we hope to have contributed to the best practices of deploying deep learning-based radiology systems at scale with findings that provide definitive insights towards advancing the use of such tools in radiology practices broadly.

\bibliographystyle{plain}
\bibliography{references}

\end{document}